\PassOptionsToPackage{colorlinks}{hyperref}
\documentclass[copyright,creativecommons]{eptcs}
\providecommand{\event}{DCM 2010} 
\usepackage{graphicx}
\usepackage{hyperref}
\usepackage[colorlinks]{hyperref}
\hypersetup{colorlinks}

\title{Computing by Means of Physics-Based Optical Neural Networks}
\author{A. Steven Younger
\institute{Center for Applied Science and Engineering \\ Missouri State University,  USA}
\email{steveyounger@missouristate.edu}
\and
Emmett Redd
\institute{Department of Physics, Astronomy, and Materials Science\\Missouri State University, USA}
\email{\quad emmettredd@missouristate.edu}
}

\def\titlerunning{Computing by Means of Physics-Based Optical Neural Networks}
\def\authorrunning{A. Steven Younger  \& Emmett Redd}
\begin{document}
\maketitle
\begin{abstract}
We report recent research on computing with biology-based neural network models by means of physics-based opto-electronic hardware. New technology provides opportunities for very-high-speed computation and uncovers problems obstructing the wide-spread use of this new capability.  The Computational Modeling community may be able to offer solutions to these cross-boundary research problems.
\end{abstract}
\thanks{This material is based on work supported by the United States National Science Foundation under Grant No. 0725867. }\

\section{Introduction}

About two decades ago, Optical Computing and Optical Neural Networks were the subjects of intense research interest.\cite{YA87}\cite{KE93}\cite{HC87} They were seen as possible solutions to the expected limits of Moore\' \,s Law. However, several problems, such as slow learning speed, high component costs, and the pushing back of the Moore\' \,s limit by more conventional technology caused interest to fade.

New technology is enabling the development of the next generation of optical computing devices. This new technology is largely driven by the explosion in optical communications. The hardware component costs have been reduced, speed has increased, and power requirements reduced. Within the past few years, very fast optical neural network learning algorithms have been developed, called Fixed Weight Learning Neural Networks (FWL-NNs).\cite{AY09}\cite{AY08}\cite{DP02}

We expect that interest in optical computing and optical neural networks will increase, especially since the Moore\' \,s Law limit seems to have been reached by conventional chip-making technology.\cite{MD05} The new optical technology should allow a speed about 10,000 times faster than a high-end laptop computer, with  projected component costs only about 10 times the cost of the laptop.

Our work is at the level of research, design, and development of prototype optical hardware computing systems and development of Fixed-Weight Learning concepts . We have uncovered some difficulties and opportunities that we believe could be addressed by the Computational Modeling community.
 
This paper reports on our work on a prototype Optical Fixed-Weight Learning Neural Network that we are developing.  We will indicate some of the new technology used in our device. Next, we will present Fixed-Weight Learning results from previous experiments on optical hardware. Finally, we describe some of the problems encountered, and point out where we believe cross-boundary, interdisciplinary research opportunities exist.

\section{Optical Neural Network Hardware}

\subsection{Optical Neural Network Prototype}

\begin{figure}
\resizebox{\textwidth}{!}
{\includegraphics {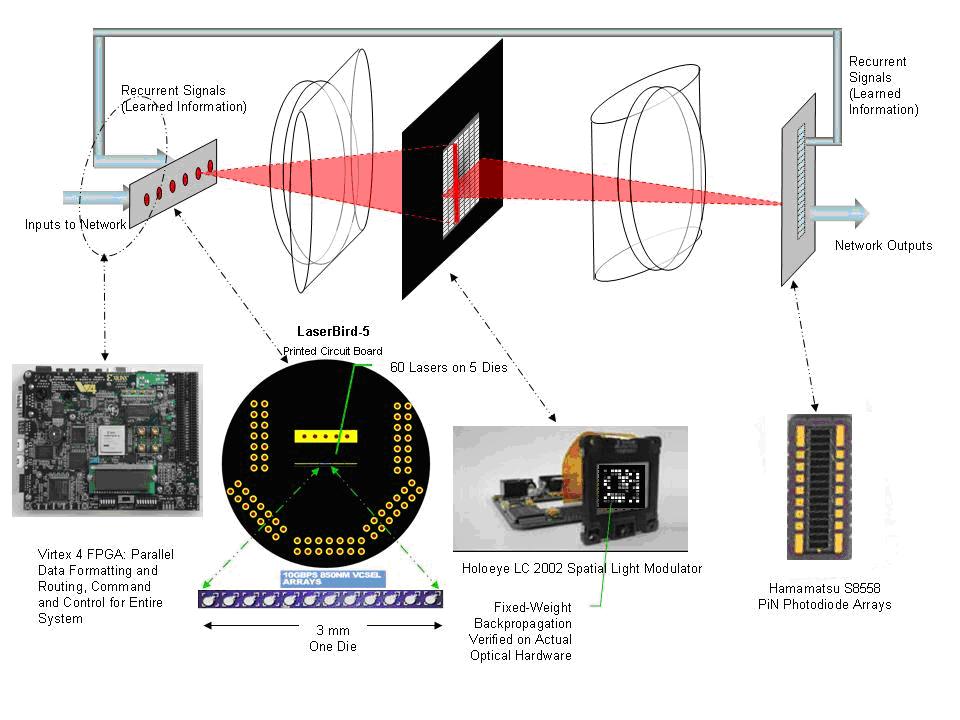}}
\caption{Optical Neural Network Prototype. Top shows a system schematic. Bottom shows some of the devices used to implement the prototype. }\label{fig:prototype}
\end{figure}

The prototype is based on the relatively old concept of a Stanford Matrix Multiplier,\cite{JG88} modified for efficient implementation of FWL-NNs (Fig.~\ref{fig:prototype}). The Stanford design multiplies an M by N matrix by an 1 by N vector argument, resulting in a M by 1 vector output.

Electronic input signals are converted to light beams by a horizontal array of laser sources (left). Note that the conversion operations are limited in precision (usually 8 to 12 bits).  However, a laser light beam itself is inherently an analog signal.  

\subsection {Optical Signal Emitters (Laser Arrays)}

Perhaps the most significant new technology is the development of low-cost, small, and very fast laser arrays. The horizontal blue bar on the bottom center left of Figure~\ref{fig:prototype} shows an linear array of 12 Vertical Cavity Surface Emitting Lasers (VSCELs), each capable of being independently modulated at 10 GHz.  The lasers are on a 3 mm long rectangular die. They can produce either binary or (usually 8 bit) analog level optical signals.  The beam from each laser is about as bright as a standard laser pointer. Each array has a potential bandwidth of 120 Gigabytes/second.  (This performance, like the estimates below, represents a best-case which is rarely achieved in a real device, mostly due to cost constraints.)

In a prototype device which we are developing, a total of 60 individual lasers are used. Future plans are to expand this to 256 sources. 

\subsection {Optical Signal Detection and Opto-Electronics}

As with sources, arrays of high-speed photodetectors have made much progress in recent years. The bottom right of Figure~\ref{fig:prototype} shows a linear array of 16 detectors based on Silicon PiN technology have reduced the cost and increased the speed.  Additionally, inexpensive operational-amplifiers, required to amplify and process the electrical output of the detectors have reached speeds in the 10 GHz range.  

The electronics side of Opto-Electronics has made similar advances in recent years. The far left of Figure~\ref{fig:prototype} shows a Virtex-4 Field Programmable Gate Array (FPGA) development board. FPGAs allow the easy creation of massively parallel digital logic devices. The FPGA configuration is specified by a Hardware Description Language, such as VHDL.  This FPGA simultaneously generates all 60 laser driver signals in our prototype device. One high-end (Virtex-6) FPGA can easily generate simultaneous driver signals for 256 VSCEL lasers and also receive signals from 256 detectors.

\subsection{Spatial Light Modulators (SLMs)}

The main computing elements of most optical computers are SLMs. They are made in a variety of forms and technologies: from Liquid Crystals and Digital Micromirror Devices to 35 mm film.  While progress has been made in SLMs (for example: million pixel devices are common), it has not been as dramatic as with the emitters and detectors.  Low-cost methods have slow update rates (or are non-changeable), and expensive devices are not much faster.  Maximum speed on most of these devices is only about 1 kHz. 

We have addressed the thinking/learning speed problem by developing a method which moves the learning into the fast signals and leaving the synaptic weights fixed.  We will discuss this in more detail below.

Physics tells us that light passing through a region on the SLM is attenuated (reduced) by an amount proportional to the gray level of the region.  We will call these attenuation gray levels weights, also denoted by ${W_{ji}}{\rm{  -  where}}$  ${i \in {\rm{Inputs,}}j \in {\rm{Outputs}}}$. The attenuation process takes only as long as the light takes to pass through the SLM. For 35 mm film, this can be as short as:
\[\Delta t = \frac{{\Delta x}}{c} \approx \frac{{{{10}^{ - 4}}{\rm{m}}}}{{3 \times {{10}^8}{\rm{m/s}}}} \approx 3 \times {10^{ - 13}}{\rm{s}}\]

A device may have 256 inputs and 256 outputs, requiring  ${256^2}$ weights. This is easily achievable on most SLMs.  The theoretical computational throughput is given by: 
\[R = \frac{{{N_{{\rm{inputs}}}} \times {N_{{\rm{outputs}}}}}}{{\Delta t}} = \frac{{{{256}^2}}}{{3 \times {{10}^{ - 13}}{\rm{s}}}} = {\rm{2}} \times {\rm{1}}{{\rm{0}}^{{\rm{17}}}}{\rm{   Operations}}/{\rm{second}}{\rm{.}}\]

We are not near this theoretical speed today. For an actual device constructed from current (2010) technology, around ${10^{13}}$  Operations/second can be achieved.  This is about 10,000 times faster than a laptop computer at a projected component cost about 10 times the cost of the laptop system.

\subsection{The Stanford Matrix Multiplier}

The Stanford Optical Matrix Multiplier design was motivated by the physics of the phenomena.  For example, the speed and relative ease of optical signal summation by cylindrical lenses.  However, current computational models have a difficult time dealing with some of the inherent features of the process, such as its hybrid analog/digital nature. 

The light from the laser array passes through a combination of lenses that spread each beam into a thin vertical line, that fall on the SLM. The beams cover the width of the SLM as a set of vertical lines (only one is shown in Fig.~\ref{fig:prototype}). Each vertical line falls on 2M horizontal segments. The parallel (analog) multiplications of the matrix product occur by the attenuation of the light beams as they pass through the SLM.  Next, the light is concentrated in the horizontal direction onto a vertical array of photodetectors. This is the means by which the summation portion of the matrix product occurs. Positive and negative values are summed by separate photodetectors (hence the 2M) and then subtracted by electronic means, usually by operational amplifiers.

A very useful extension of this model is the (artificial) neural network, a model that was inspired by biological neural networks. 

To convert the Stanford Multiplier hardware into neural network hardware merely requires that electro-optical signals be able to drive the operational amplifier into the saturation portion of its output range. The "natural" saturation curve of the amplifier serves as the nonlinear neural squashing function.

Optical neural networks based on the above model are very fast in performing the forward propagation, or \emph{thinking} phase of their operation.  However, one of the strong reasons for using neural networks it that they have the ability to learn computational tasks (function mappings) from training data. This is particularly useful for tasks where the algorithm is unknown.

Traditional optical neural schemes are slow during the neural learning phase because they require changing a relatively slow synaptic media, such as a Spatial Light Modulator (SLM). This requires milliseconds vs.  nanoseconds for the thinking phase.

\subsection{Optical Encoding of Neural Signals}

At a given time, each neuron has an output signal or activation.  The method of encoding or representing these signals is an important issue, especially in hardware-based systems. In biology, the signals are mainly in the form of pulse trains. We know that the average pulse rate is important to the behavior of the network. However, the importance of other features, such as pulse shape, variability of inter-pulse timing, and neuron activation history are open questions at this time.  Most artificial neural models use only pulse rates to encode neural activations. In a digital computer, these pulse rates are often represented by a floating-point number.  In optical systems, this rate must be encoded in an analog signal.

There are a variety of ways to optically encode activations. The brightness or intensity of the laser light is probably the fastest, but it requires fairly complex interface and drive electronics. The precision of the signal is limited to some number of bits (usually 8 to 12).  The method that we mostly use, called Stochastic Pulse encoding (SP), more closely mimics biological neural signals. It also uses simpler electronics. Its main drawback is that it is slightly slower than intensity encoding. 

\begin{figure}
\resizebox{\textwidth}{!}
{\includegraphics {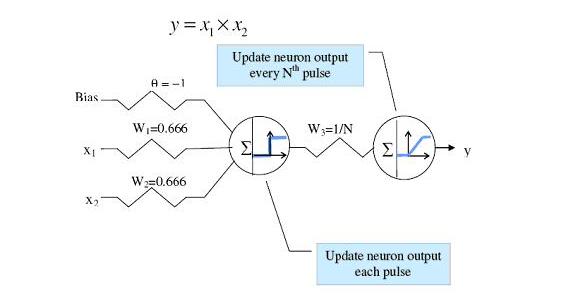}}
\caption{ $\Sigma  -  \wedge$ (Sigma-And): First-Order Neural Network to multiply a-signal-times-a-signal . }\label{fig:sp}

\end{figure}

In SP encoding, the optical signal interval is divided into a number of time slices (usually 256). At each time slice, the signal is either a 1 or a 0. The probability of a 1 for any given time slice is proportional to the neural activation level. These networks are slower because a number of pulses (usually 256) are used to represent an activation.  Figure~\ref{fig:sp} shows a very efficient way to multiply a Signal-times-a-Signal.\cite{SB94} When two statistically independent signals are combined with an AND operation, the probability of both signals being a 1 for any given time slice is equal to the product of the input signals. It is hard to imagine a "cheaper" way to compute a product -- averaging the output of an AND gate over a number of pulses. 

We have modified the standard SP to also allow attenuated pulses instead of being restricted to unit pulses. We developed small neural networks to carry out the Signal-times-a-Signal operation, which we called $\Sigma  -  \wedge $  (Sigma-And) networks.

\section{Fixed-Weight Learning}
\subsection{What is it and how does it work?}
FWL-NNs learn new mappings by dynamically changing recurrent neural signals. The (fixed) synaptic weights of the FWL-NN implement a learning algorithm which adjusts the recurrent signals toward their proper values. Instead of encoding a particular mapping, the synaptic weights of a FWL-NN encode how to learn any mapping (within a large, perhaps infinite, set of possible mappings).  Corrective feedback maintains the value of the recurrent signals during learning.

Fixed-Weight Learning is somewhat analogous to biological working memory, where information storage is also believed to involve a persistent pattern of activated neurons maintained by feedback.

FWL-NNs can be implemented by many means, such as pure software, digital electronic hardware, or analog electronic hardware. FWL-NNs are a particularly good match for the capabilities given by optical neural networks, since it replaces slow SLM updates with high-speed changes in laser light intensity.  Also, FWL-NNs can learn using the slowest and least expensive synaptic media, 35mm film. 

Of course, the FWL-NN is (usually) larger for a given problem than a changing-weight network, because it contains its own learning algorithm.  Given the high density of optical media, this trade-off is often acceptable. 

In 1991, Cotter and Conwell \cite{NC91} presented and proved the Fixed-Weight Learning Theorem:  Given a neural network topology (which learns by changing weights) and its attendant learning algorithm, there exists an equivalent FWL-NN. Any mapping that can be learned by changing the weights of the original network can be learned by the FWL-NN without changing any synaptic weights. Note that the FWL-NN is necessarily recurrent, even if the equivalent original changing-weight network was not.

The Fixed-Weight Learning Theorem is an existence proof based on Universal Approximation. It did not specify how to generate the FWL-NN.  Several methods have been used by us and other researchers \cite{AY09}\cite{AY08}\cite{DP02} to derive the fixed synaptic weights. 

We have designed, and verified on actual optical hardware several FWL-NNs based on the Backpropagation of Errors learning algorithm.  Some results are described below. 

\subsection{Training the Fixed-Weights}
As noted above, in a FWL-NN the synaptic weights encode a learning algorithm (e.g., Backpropagation).  The (fixed) synaptic weights that implement the learning algorithm were derived by a procedure we call the \emph{sub-network} method.  First, break the learning algorithm to be encoded (for instance, Backpropagation) into smaller sub-tasks. Next, train small, feed-forward sub-networks to perform these sub-tasks. (Some sub-networks are simple enough to be designed manually.) Finally, integrate the sub-networks together to form the complete FWL-NN.

During the integration process, care must be given to providing the proper signal delays for the various sub-networks. For this research, we performed the integration manually, which was time-consuming and prone to errors. 

\section{Hardware-Based Experimental Results}

We designed and tested several FWL-NNs on an Optical Hardware Test Bench, described in \cite{AY09}\cite{AY08}. The test bench performed the synaptic multiplications by the optical attenuation process. However, it performed the summation, integration, and squashing functions in software.  The synaptic multiplications are the vast majority of the computations for a neural network, and are usually the most problematic portion of the optical calculations.

All of the FWL-NNs presented below perform \emph{on-line learning}, where the network is constantly adapting its mapping based on the feedback information. However, this is not a limitation inherent in FWL-NN in general. 

Three of these networks are described in Table~\ref{tab:networks}. The number of layers, \begin{table} 
\caption{OPTICAL FIXED-WEIGHT NEURAL NETWORKS. Name and number of layers, neurons, and synapses in the network. Generic encoding networks will work with almost any activation signal encoding scheme. The  $\Sigma-\wedge$  encoding must use Stochastic Pulse Modulation.}\label{tab:networks}
\begin{center}
\begin{tabular}{|l|l|l|l|l|}
\hline
\textbf{NN}&\textbf{Layers}&\textbf{Neurons}&\textbf{Synapses}&\textbf{Encoding}\\
\hline
\emph{uMULT}&3&13&30&Generic\\
\hline
\emph{PlanTran}&4&29&100&Generic\\
\hline
\emph{BooLean}&5&33&56&$\Sigma-\wedge$\\
\hline
\end{tabular}
\end{center}
\end{table}
neurons, and synapses are given. The “Encoding” column indicates the type of activation level signal encoding that the network requires. \emph{Generic} means that almost any encoding method (Intensity-Based, Pulse-Width-Modulation, uniform pulse trains, and Stochastic Pulse) can be used with the network.  The $\Sigma  -  \wedge$ means that only the modified Stochastic Pulse method described above can be used.  The compactness of the $\Sigma  -  \wedge$  network can be appreciated by noting that the Generic version of the BooLean network (not reported here)  has over 300 synapses.

The first network, \emph{uMULT}, is a Feed-Forward neural network (not a FWL-NN). It is an example of a sub-network that performs the “signal-times-a-signal” operation using generic signal encoding.  It has 2 inputs and one output value.  As can be seen in Table~\ref{tab:performance}, \begin{table} 
\caption{ PERFORMANCE OF NETWORKS ON HARDWARE.  \textbf{${N_P}$} is the length of the pulse train, \textbf{${N_C}$} is the number of cycles required for network convergence (learning time), and \textbf{MSE} is the mean-squared-error AFTER learning has occurred. }\label{tab:performance}
\begin{center}
\begin{tabular}{|l|l|l|l|}
\hline
\textbf{NN}&\textbf{${N_P}$}&\textbf{${N_C}$}&\textbf{MSE}\\
\hline
\emph{uMULT}&128&n/applicable&0.0013\\
\hline
\emph{PlanTran}&256&13&0.0083\\
\hline
\emph{BooLean}&256&15&0.0140\\
\hline
\emph{BooLean}&256&21&0.0076\\
\hline
\emph{BooLean}&1024&75&0.0098\\
\hline
\end{tabular}
\end{center}
\end{table}
it can perform the mapping with good accuracy (5 bits output with 7 bits input). ${N_P}$  is the number of pulses in the signal activation pulse train.  Note it is also much larger than the   $\Sigma  -  \wedge$  network (Fig.~\ref{fig:sp}).  \emph{uMult} has 13 neurons and 30 synapses vs. the  $\Sigma  -  \wedge$  network which performs the same mapping with only 2 neurons and 5 synapses.

\emph{PlanTran} is a FWL-NN that uses Generic encoding. It (theoretically) can learn any function mapping that can be generated by a single-synapse sigmoid neuron network. That is the mapping:

\[\begin{array}{l}
 y = {\mathop{\rm logsig}\nolimits} \left( {W \times x} \right),  {\rm{ where}}\\
x,y \in [0,1),{\rm{ }}W \in [ - 4, + 4],{\rm{ }}{\mathop{\rm  logsig}\nolimits} \left( s \right) = {\left[ {1 + {e^{ - s}}} \right]^{ - 1}} \\ 
 \end{array}\]

The test data for \emph{PlanTran} was generated by first randomly selecting a \emph{W}. A large number of data pairs are generated by repeatedly selecting a random \emph{x} and computing a \emph{y} based on the \emph{W} and \emph{x} values.  In order to learn the mapping, \emph{PlanTran} must derive \emph{W} from the data. Recurrent feedback loops, along with the data values from the previous step, tell \emph{PlanTran} (via its embedded learning algorithm) what it needs to do to improve its performance. 

\emph{PlanTran} has an on-line version of Backpropagation of Errors algorithm embedded or encoded in its synapses.  The  ${N_C}$ value in Table~\ref{tab:performance} is the average number of steps that the FWL-NN took to converge. That is, how long the FWL-NN took to learn the function mapping.  The decision as to when learning has occurred is somewhat subjective, since the FWL-NN will have small errors even after convergence (third column). However, we believe that any reasonable decision rule will lead to similar results. 

The rates shown in Table~\ref{tab:performance} are very comparable to those for non-fixed-weight on-line Backpropagation. 

\emph{BooLean} is a FWL-NN capable of learning any linearly separable function mapping with two Boolean arguments and one Boolean result. There are 14 such functions.  For instance, the network can learn to be an “AND” gate, an “NOR” gate, or whatever type of gate it needs to be in order to reduce the error. It also has standard Backpropagation learning encoded in its fixed synaptic weights. Functionally, BooLean is essentially three of the PlanTran networks combined.  However, we used the  $\Sigma  -  \wedge$  network topology because of its much smaller size. Note that, because of this, \emph{BooLean} is only about half the size of \emph{PlanTran} although it can learn mappings that are about three times harder.

There are three examples in Table~\ref{tab:performance}, including one with 10-bit activation signals, which took longer to converge and even performed slightly worse than the 8-bit examples. This is an indication that random errors are probably not the factor limiting network precision. 

\begin{figure}
\resizebox{\textwidth}{!}
{\includegraphics {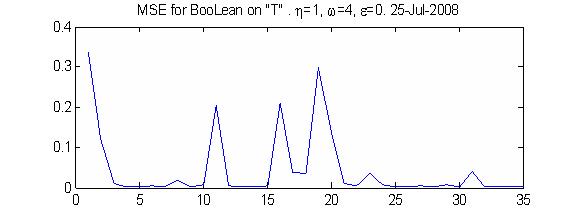}}
\caption{Mean Squared Error of  FWLNN\emph{ BooLean}.  The network required about 21 steps to learn the mapping. }\label{fig:mse}
\end{figure}

Figure~\ref{fig:mse} shows the plot of \emph{BooLean} learning the function mapping \emph{Always True}. The network converges at about step 21. Small residual errors after learning are also shown.

\section{Building/Verifying Hypercomputational Hardware}

The above described hardware is a mixture of analog and digital components.  We have had a long-term interest in moving more of the computing into the optical path.  This will also make more of the calculations be analog.  Ultimately, we may succeed in building a completely analog recurrent neural network (ARNN) which can perform hypercomputation.\cite{HS99}  Related to this hypercomputation may be the realization of a true Universal Approximation neural network (UANN).\cite{JW94}

However, this is real, physical hardware which contains noise that will ultimately limit precision.  Although Siegelman \cite{HS99} claims “linear precision suffices,” noise may thwart being able to 
build a P/poly computation machine and limit the machine to BPP/log.\cite{HS03}  Noise may also thwart being able to make a true UANN.  In either case, our expertise lies in building hardware.  We seek collaboration with computation modelers for help on several activities: 	
\begin{quote}
\begin{tabular}{l}
$\bullet$ Directing our hardware development efforts to best achieve hypercomputation,\\
$\bullet$ Designing problems which can verify its true computational power, and\\
$\bullet$ Seeking funding for studies attempting to develop true artificial intelligence.\\
\end{tabular}
\end{quote}
We appeal to the Computational Modeling community to help us build computational machines that are truly smarter, not just faster.

\bibliographystyle{eptcs} 

\end{document}